\documentclass[12pt, letterpaper]{article}
\usepackage{color}
\usepackage{gensymb}
\usepackage{graphicx}
\usepackage{caption}
\usepackage{subcaption}
\usepackage{multirow}
\usepackage{setspace}
\usepackage{url}

\newcommand{\AP}{\textbf{AP}}
\newcommand{\APR}{\textbf{AP-Refine}}
\newcommand{\LP}{\textbf{LP}}
\newcommand{\LLP}{\textbf{LLP}}
\newcommand{\APLP}{\textbf{APLP}}
\newcommand{\APLLP}{\textbf{APLLP}}
\newcommand{\KM}{\textbf{KMeans}}
\newcommand{\Full}{\textbf{Full}}
\newcommand{\Mean}{\textbf{Mean}}
\graphicspath{{./imgs/}}

\providecommand{\keywords}[1]{\textbf{\textit{Keywords---}} #1}

\title{A Rapidly Deployable Classification System using Visual Data for the Application of Precision Weed Management}
\author{David Hall\thanks{The authors are with the School of Electrical Engineering and Computer Science, Queensland University of Technology (QUT), Brisbane, Australia email: d11.hall@hdr.qut.edu.au, tristan.perez@qut.edu.au and c.mccool@qut.edu.au}, Feras Dayoub\thanks{The author is with the ARC Centre of Excellence for Robotic Vision, Queensland University of Technology (QUT), Brisbane, Australia. http://www.roboticvision.org/ email: feras.dayoub@qut.edu.au}, Tristan Perez\footnotemark[1] and Chris McCool\footnotemark[1]
}
\begin{document}

\maketitle
\begin{abstract}
	In this work we demonstrate a rapidly deployable weed classification system that uses visual data to enable autonomous precision weeding without making prior assumptions about which weed species are present in a given field.
	Previous work in this area relies on having prior knowledge of the weed species present in the field.
	This assumption cannot always hold true for every field, and thus limits the use of weed classification systems based on this assumption.
	In this work, we obviate this assumption and introduce a rapidly deployable approach able to operate on any field without any weed species assumptions prior to deployment.
	We present a three stage pipeline for the implementation of our weed classification system consisting of initial field surveillance, offline processing and selective labelling, and automated precision weeding.
	The key characteristic of our approach is the combination of plant clustering and selective labelling which is what enables our system to operate without prior weed species knowledge.
	Testing using field data we are able to label 12.3 times fewer images than traditional full labelling whilst reducing classification accuracy by only 14\%.
\end{abstract}


\section{Introduction}\label{sec:intro}

\begin{figure}[t]
    \centering
    \includegraphics{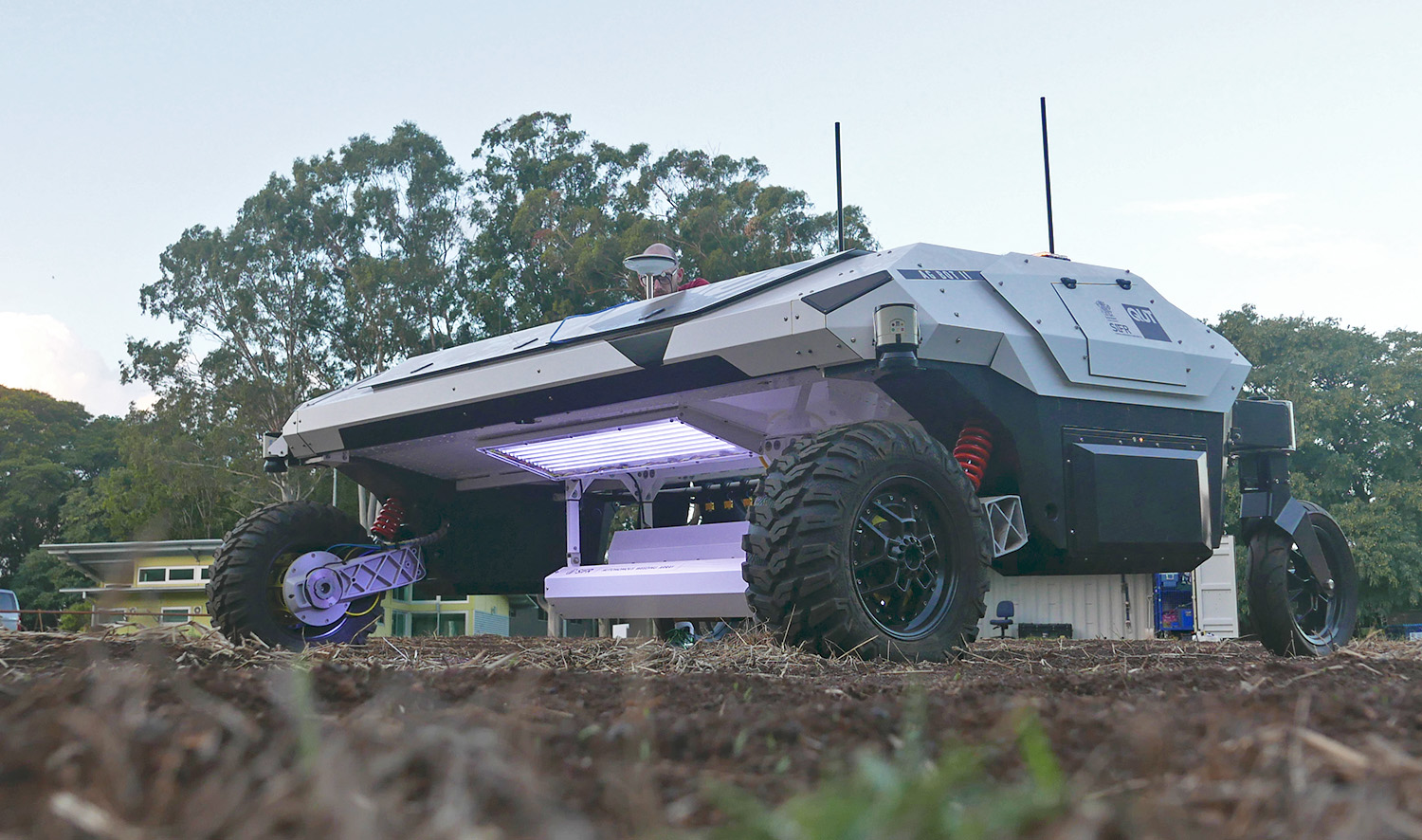}
    \caption{AgBot II agricultural robotics platform}
    \label{fig:intro:agbot}
\end{figure}

Farmers have seen a steady increase in herbicide resistance from various species of weeds over the years~\cite{Gilbert2013}.
This has led to an increased research focus on precision weed management strategies where each weed is treated individually using a treatment which best suits that plant.
To do so manually is laborious and costly when covering large areas.
This, in combination with other factors, has led to a growing interest in the potential of agricultural robotics capable of performing autonomous precision weeding such as the AgBot II shown in Figure~\ref{fig:intro:agbot}.

Achieving autonomous precision weeding has been a focus of research for many years yet weed classification still remains a critical problem which is generally considered unsolved within this field~\cite{Slaughter2008}.
Weed classification is usually done using either species-specific classification~\cite{Gerhards2006} or weed-vs-crop classification~\cite{Aastrand2002}.
While successful for their  specifically targeted tasks, these techniques either assume that they know exactly what species are present in the field or assume that species-specific knowledge is not necessary.
Neither assumption is valid if we want species-specific weed management which is easily deployable to any given field.
To the best of our knowledge, the challenge of creating rapidly deployable weed management systems that can perform intra-row weeding without needing any prior training has only been considered a few times before and these only consider weed-vs-crop classification systems~\cite{DeRainville2012}~\cite{Strothmann2017}~\cite{Wendel2016}.
Neither one tackles the challenge of providing a species-specific weed management system which can be deployed rapidly in a field without knowing the species within in advance.
This challenge of rapidly deployable species-specific precision weeding is what our research works to enable.

The main contribution of this work is the creation and demonstration of a rapidly deployable species-specific weed classification system for use in autonomous precision weeding.
This system is able to operate without assuming that it knows what weed species are present prior to deployment and can be trained on a specific field with minimal human effort.
The system is novel in many ways from what is currently found in autonomous weed classification literature.
Unlike the current literature we:
\begin{itemize}
    \item Use unsupervised clustering to summarise weed species
    \item Use selective labelling based on the clustering results to fully label scouted plant data rapidly with minimal human effort
    \item Use the data labelled from selective labelling to train classifiers able to operate in the given field after one scouting operation
    \item Use highly descriptive learnt features with low dimensionality (128 vs 1024 dimensions) to improve clustering results.
\end{itemize}
We evaluate several methods for clustering and selective labelling for our classification system to evaluate which approaches work best for our task.
We show that on a dataset of weed images collected in a field we could label 12.3 times fewer images than full labelling and still achieve labelling accuracy of 79\% and a decrease in classification accuracy of only 14\% from full labelling.
Through our analysis we identify the strengths and weaknesses of the different methods tested and propose directions for future work.

\subsection{Literature Review}\label{sec:intro:lit}
There have been several classification approaches for autonomous weeding presented over the years, using a variety of plant descriptors and classifiers.
These approaches try to solve either a binary weed-vs-crop, or a multi-class species-specific classification task.
Both approaches tend to use either shape, reflectance, or texture features - and often combinations of thereof~\cite{Slaughter2008}.

Weed-vs-crop classification methods can be seen as the simpler of the two options, being a more easily implemented weeding procedure for specific crops as it considers only a two class problem and can take advantage of the structured manner that crops are usually sown.
One technique used in weed-vs-crop classification is to perform only inter-row weeding without any need for a specific plant-wise classification.
Such a technique was by Emmi et al. where the main focus was on calculating weed densities between the crop rows in order to utilise resources more efficiently~\cite{Emmi2014}.
This simplifies the classification problem by only needing to detect plants and crop rows rather than needing to distinguish a specific species within the detected plants.
In some applications, the weed-vs-crop classification task is made easier by the size difference between crop and weed~\cite{Blasco2002}.
Blasco et al. focused on cabbage crops which are a transplanted crop and most likely always bigger than the weeds around them~\cite{Blasco2002}.
This allowed for a simple size threshold to identify crop and weed.

Other works deal with a harder crop-vs-weed plant-wise classification procedure with crops of similar size to the weeds~\cite{Aastrand2002}~\cite{Lottes2017}.
This harder process for classification can however be aided by the use of positional information.
This takes advantage of known prior information about how the crop should be planted such as knowing that the crop is sown in rows and with an expected spacing between them.
Lottes et al. calculated distances between query objects or keypoints and plants previously classified as crops~\cite{Lottes2017}.
These distances, combined with other information are used to calculate the probability that a given plant is a crop based on these distances.
This probability was then used as a feature within their classification system.

Weed-vs-crop methods which utilise plant-wise classification are typically only designed for a single crop and would require retraining before they could be used for different scenarios, which can be a time-consuming process as crop image data needs to be collected and fully manually annotated.
While these crop-vs weed systems are undoubtedly useful and easily implementable they do have one major drawback which is that they treat all weeds as the same.
This does not allow for species-specific treatment which may be required by some farmers.

In order to achieve species-specific treatment, a multi-class classification approach is required.
Lin demonstrated the use of an SVM classifier with shape features to achieve 75.00\% and 82.85\% average classification accuracy in field and greenhouse tests respectively across 6 pre-defined species~\cite{Lin2009}.
The precision weeding system presented by Bawden et al. demonstrated a classifier trained to identify five different weed species which successfully identified up to 98.8\% of one of the classes correctly.
While impressive, the system did struggle distinguishing between different grass species due to the large visual similarity between such plants, in one case identifying only 47.5\% examples of one species correctly~\cite{Bawden2017}.
A unique system was developed by Haug et al for classifying overlapping plants by classifying a grid of small patches across segmented plant regions and then interpolating the results of each patch until whole plant regions were classified, achieving an accuracy of 93.8\%~\cite{Haug2014}.
It should be noted that in this case, of the three defined classes, only two were defined plant species and the third class was simply labelled as ``other weeds''.

The use of a single class to define all other weed species found in a field is not uncommon within this field of research. 
Sometimes this ``other'' class is split into ``other grass'' and ``other broadleaf'' classes such as was done by Gerhards and Obel~\cite{Gerhards2006}.
In that work, Gerhards and Obel classified three different weed species as well as ``other broadleaf'' and ``grass weeds'' and tested the system in the field.
This system managed to provide herbicide reductions of up to  81\% and increased weeding efficacy between 85\% and 98\% using this classification system.
In a more recent work, Lottes et al. evaluated both a weed-vs-crop classification system as well as a species-wise plant classification system with three pre-defined plant species as well as an ``other weeds'' class~\cite{Lottes2017}.
Using shape, reflectance, texture and position features they achieved an overall accuracy of 86\% of predicted objects for their species-wise system and  96\% accuracy for crop vs weed classification.
The main detractor for the precision of their system was stated as being due to the performance of the ``other weeds'' class.
This was hypothesised as being because this class has a small number of samples and a high intra-class variance as it represented every other weed species not previously defined.
This use of an ``other weeds'' class highlights a problem inherent to classification approaches for automated weed management.
This problem is that they need prior information about which species are to be expected and cannot be adapted for different species if they are transferred to fields which do not meet with the prior assumptions being made.
The algorithms need previously created, manually labelled datasets in order to be retrained which can be a laborious and slow process which won't necessarily meet with a farmer's immediate needs. 

One of the few works that approaches the weed classification task without making prior assumptions about the plant species in the field is work done by DeRainville et al~\cite{DeRainville2012}. 
They performed intrarow weed-vs-crop weed management where the crop species was not predefined.
The only assumption made was that the crop was planted in rows. 
All plants not within the crop rows were considered to be weeds. 
Using this assumption they could infer a model for the weeds and crop and achieved a crop classification accuracy on average of 94\% across two different crops. 
A limitation of this work is that it is only applicable once the crop has grown to a state where crop rows can be detected.
This means that it cannot be applied in the fallow period before the crop has begun to grow or when the crop is young enough that the crop rows are not easily detected.

Wendel and Underwood~\cite{Wendel2016} had a similar approach to DeRainville by taking advantage of crop rows to generate training data for a crop vs weed classification system.
After crop rows are detected and plant regions extracted, any plant region which does not touch any crop row is considered as a weed plant and all pixels within said reason are labelled as being weed pixels.
Crop pixels were taken as any plant pixel within a one pixel thick line running through the center of the detected crop row.
While not all of these pixels are guaranteed to be crop pixels the majority are assumed to be so.
These automatically labelled pixels are then used to train a classifier using hyperspectral information to perform the final crop vs weed classification.
This work is limited in the same manner as DeRainville as the crop rows are necessary for the algorithm to operate, however unlike DeRainville, this system can operate in a pixelwise fashion and is better suited to overlapping plants.

Aside from DeRainville and Wendel and Underwood, Strothmann et al.~\cite{Strothmann2017} is the only other work we have found which performs autonomous weeding without making assumptions regarding what plant species are to be expected in advance.
Here, a small portion of the field is sampled using a high resolution sensor array on an agricultural robot and then the resulting image is labelled in a pixelwise fashion by a human user.
The labelled data is then used to train a pixelwise weed-vs-crop classifier for precision weeding.
They test the system across two different crops and show great accuracy increases by retraining their system for each crop, reducing critical error between 79.7 and 100\%.
This system, however, also has limitations to widespread adoption of such a strategy.
The precise sensor technology used to generate their high resolution images requires the robot to travel at a top tested speed of 0.4 m/s.
This is too slow for many broadacre agricultural robotics tasks.
All three methods described also only consider a weed-vs-crop classification scenario which removes the possibility of implementing a species-specific weeding strategy using their approaches.
While the system presented by Strothmann could be adapted for this task, there is no guarantee that the small sample area covered for labelling would contain all species present in the field.
Covering the entire field and then manually searching for each plant species within a full field image would also be incredibly time consuming.

An alternative approach for species-specific automated weeding without prior weed species assumptions had been presented in our previous work~\cite{Hall2017}~\cite{Hall2017a}.
These introduced the idea of using unsupervised clustering to summarise weed species data~\cite{Hall2017} and selective labelling of data to train a final classifier to be used for an automated precision weed system~\cite{Hall2017a}.
It is the principles in these works which we build upon for our final proposed rapidly deployable weed classification system for species-specific automated precision weeding.
\section{Methodology}\label{sec:method}

\begin{figure}[t]
    \centering
    \includegraphics[width=0.95\textwidth]{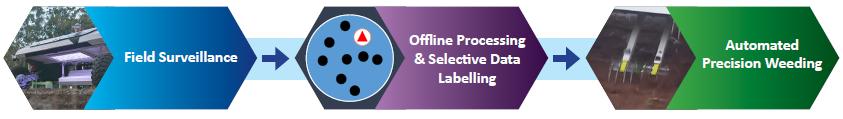}
    \caption{Main pipeline of proposed adaptable precision weeding process}
    \label{fig:method:pipeline:main}
\end{figure}

Expanding upon our previous studies~\cite{Hall2017}~\cite{Hall2017a}, this work demonstrates a rapidly deployable classification system for weed management by agricultural robotics.
We propose a pipeline for utilising this classification system within an automated precision weed management system, enabling the weed management to operate without any prior weed species knowledge.
This pipeline is shown visually in Figure~\ref{fig:method:pipeline:main} and consists of three main stages.
The main stages of our pipeline are field surveillance/data collection, offline processing and selective data labelling, and automated precision weeding.
These stages and the procedures utilised within each stage is outlined in further detail in the following sections.

\subsection{Field Surveillance}
\begin{figure}[t]
    \centering
    \includegraphics[width=0.95\textwidth]{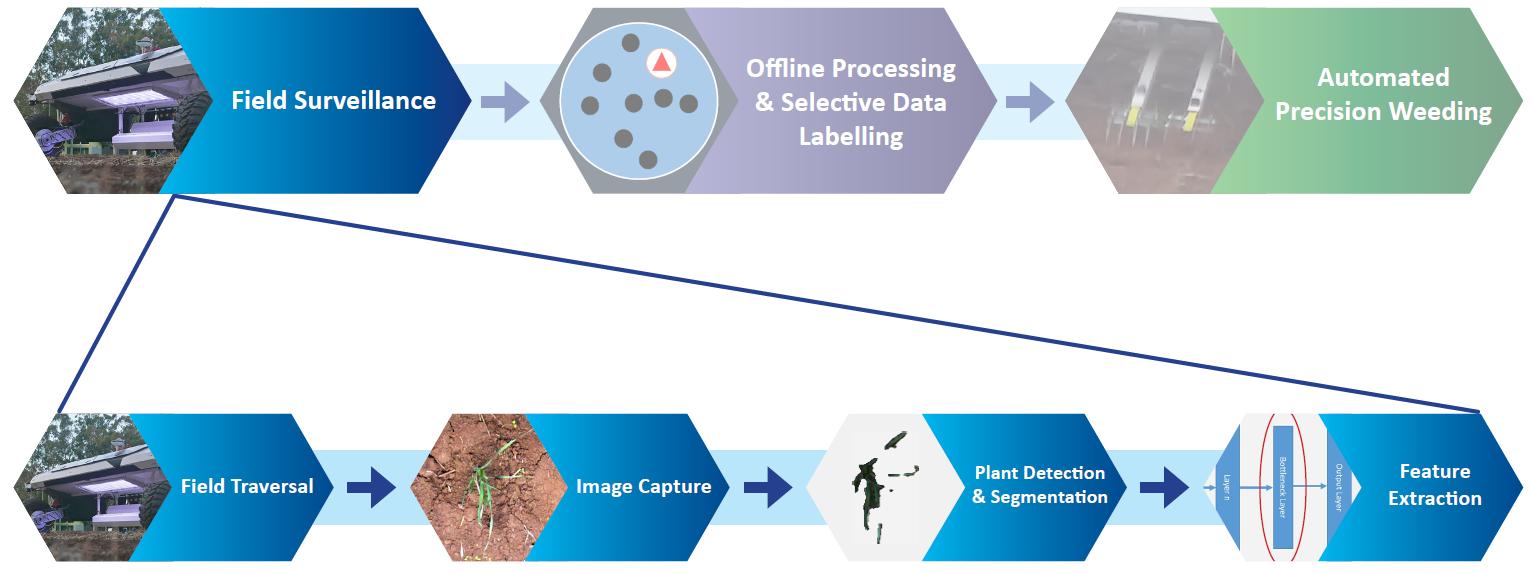}
    \caption{Field surveillance/data collection pipeline from stage one of main pipeline}
    \label{fig:method:pipeline:survey}
\end{figure}

The first stage of our pipeline is initial field surveillance.
The procedure for this stage is shown in Figure~\ref{fig:method:pipeline:survey}.
The stages here are defined as field traversal, image capture, plant detection and segmentation, and feature extraction.
The data and features collected in this stage are what is used during offline processing and selective data labelling stage of the main pipeline.
In this work we do not focus on the field traversal and image capture stages of the pipeline as these should be a part of any given agricultural robots operating procedure and are a matter of hardware and not data processing.
Those challenges are outside the scope of this work.
The following two sections shall provide more detail on the plant detection and segmentation and feature extraction processes utilised within this work.

\subsubsection{Plant Detection and Segmentation}\label{sec:method:detect}
\begin{figure}[t]
    \centering
    \includegraphics[width=0.8\linewidth]{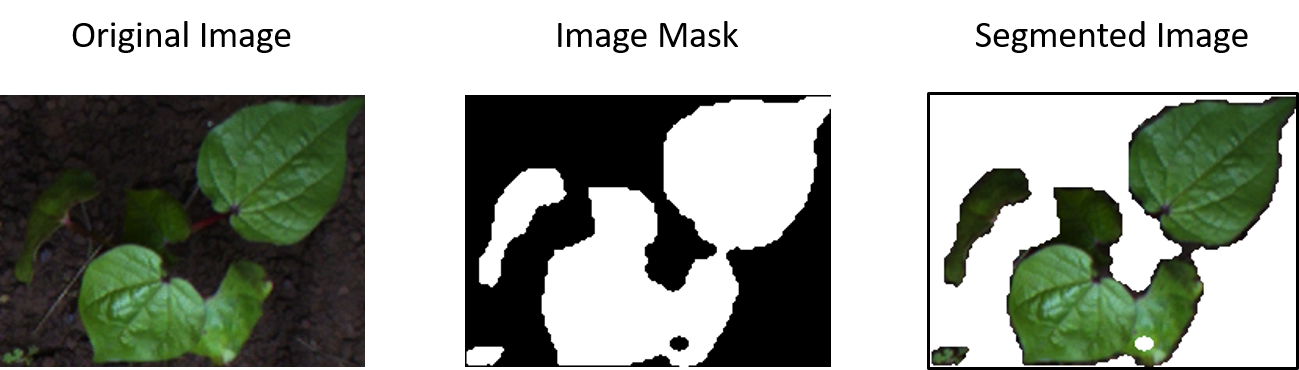}
    \caption{Example of plant segmentation with a mask being generated of the original image and then being applied to said image to present the final segmented image used for feature extraction and final classification. Note that the background of the segmented image has been changed from black to white for visualisation purposes. Best viewed in colour.}
    \label{fig:method:detect:seg}
\end{figure}

For detecting and segmenting plants using an agricultural robotic platform, we use the technique outlined in detail within previous work~\cite{Hall2017, Bawden2017}.
Plant pixels are segmented using a multivariate Gaussian classifier trained on illumination invariant colour channels from three different colour spaces.
This multivariate Gaussian is trained on a dataset which is different from any dataset used in this work for evaluating our proposed pipeline.
Segmentation is tidied up and smaller detections filtered out and then the segmentation mask is applied to the original image providing a final segmented plant image such as is shown in Figure~\ref{fig:method:detect:seg}.
As a robot traverses a field, multiple images may be captured of a single plant.
Some algorithms that we test in this work, outlined further in Section~\ref{sec:method:clust} take advantage of these multiple observations and so, in the detection stage each observation of a single plant should be appropriately recorded.
We do not focus on this procedure in this work and so perform this manually for our experiments, however, such a task can be accomplished with plant localisation using techniques such as the use of highly accurate global positioning systems (GPS).
Within our pipeline, once plants have been detected and segmented from their background, we can extract features to enable the grouping of similar plant images together.

\subsubsection{Feature Extraction}\label{sec:method:feat}
A key component for any plant recognition approach is the use of features to describe the physical characteristics of individual plants.
In our pipeline, we extract learnt features which enable our plant clustering algorithms to group plants with similar physical characteristics together without prior knowledge of what each plant group is expected to be.
The learnt features used here are deep convolutional neural network (DCNN) descriptors.
In previous work, we had already established that DCNN descriptors could achieve higher baseline classification accuracy than hand-crafted features (HCFs) more typically seen within the literature for the task of leaf classification~\cite{Hall2015}.
In this work, we use the same bottleneck DCNN features outlined in~\cite{Hall2017} that were proved to be the most effective descriptors for unsupervised weed clustering.

These bottleneck features are extracted from an adjusted version of the 22 layer deep inception DCNN architecture, more commonly referred to as GoogLeNet~\cite{Szegedy2015}.
Our adjustment to this architecture, is the addition of a bottleneck layer which is used to extract low-dimensional descriptive features.
The bottleneck layer is a fully connected layer with fewer neurons than the layer which preceded it and the descriptor used is the output of each neuron of this layer.
In our case, the bottleneck layer consists of 128 neurons providing a final 128 dimensional descriptor vector for each plant image.
This bottleneck layer is critical as it reduces the dimensionality of the resultant descriptor vector from 1024 to 128.
In previous work~\cite{Hall2017} we established that these lower dimensional features overcome the ``curse of dimensionality'' which can affect simple clustering algorithms.
The bottleneck features were found to improve clustering accuracy for most of the algorithms tested.


Following our previous procedure~\cite{Hall2017}, we fine-tune the bottleneck DCNN network on plant images to improve the network's ability to discriminate between different observed weed species by refining the scope of the images that the network is trained on.
This presents its own difficulties as, in order to provide unbiased testing, fine-tuning data should not contain the same classes as are used for testing.
To this end, we fine-tune our network on leaf training images from the 2016 PlantCLEF challenge~\cite{Goeau2016} which is a large but unrelated plant dataset with images taken in a different setting than either of our testing datasets.
While these images differ to those in either of the testing datasets, they are closer to the scope of our task than the images which had originally been used for training the network.
Fine-tuning the network on the PlantCLEF data therefore improves features attained due to this focus in scope.
Finally, we perform L$_2$ normalisation on the descriptor vector that we get as output from the bottleneck layer before using it in other stages of our pipeline.



The feature extraction stage is the final part of the field surveillance stage of our main pipeline.
After the full field has been traversed and features are extracted for each detected and segmented plant, we enter the offline processing and selective data labelling stage of our main pipeline.

\subsection{Offline Processing and Selective Data Labelling}\label{sec:method:op_and_sd}
\begin{figure}[t]
    \centering
    \includegraphics[width=0.95\textwidth]{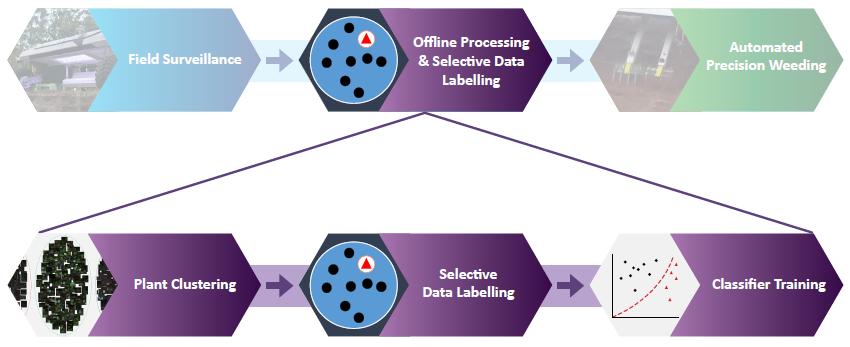}
    \caption{Offline processing and selective data labelling pipeline of stage two of main pipeline}
    \label{fig:method:pipeline:offline}
\end{figure}

The main focus of our work is the second stage of our pipeline, the offline processing and selective labelling of data.
This is what enables us to rapidly specialise our precision weeding system to a given field without advance knowledge about weed species.
The pipeline for the offline processing and selective labelling stage is shown in Figure~\ref{fig:method:pipeline:offline}.
Using the information collected in the field surveillance stage, this procedure consists of plant clustering, selective data labelling, and training a final weed classifier.
Here, we group plants which appear visually similar together, select a few representative plant images (exemplars) for each group to be labelled by a human user, use these few labelled plants to label all plants seen in the field surveillance stage, and finally train a classifier which is able to operate within the field without further supervision.

In this work we present and evaluate multiple strategies for clustering followed by selective data labelling as these are what enable our rapidly deployable classification system to operate without needing to predefine weed species.
In the following sections we shall outline the plant clustering, selective data labelling and classification systems used within this work.

\subsubsection{Plant Clustering}\label{sec:method:clust}
The first stage of our offline processing and selective labelling pipeline is the unsupervised clustering of all weeds into visually similar groups.
This acts as an aid to data summarisation and allows for quick data labelling by farmers.
Unsupervised weed clustering is a key aspect of our approach in that it groups plants together without any need to know what the plants are in advance.
It was the first stage investigated in our previous work~\cite{Hall2017} and is critical for enabling rapidly deployable precision weeding.
In this work, we examine the use of three different clustering algorithms for our analysis into improved clustering and data labelling combinations.
The three different clustering algorithms used in this work are affinity propagation (AP)~\cite{Frey2006}, locked agglomerative hierarchical clustering~\cite{Hall2017}, and k-means clustering~\cite{MacQueen1967}.

\vskip 3mm

\noindent \textbf{Affinity Propagation}

\noindent Affinity propagation (AP) clustering is a relatively recent clustering algorithm developed by~\cite{Frey2006}.
It operates via an iterative message passing process between each sample provided to the algorithm.
The aim of this algorithm is to find exemplar samples which are representative of a single cluster of samples.
The procedure involves two types of message being passed between samples.
Firstly each sample sends a message of responsibility to each other sample (candidate exemplar) to reflect how well these candidates are suited to be an exemplar for this sample.
This is initialised using a similarity matrix depicting how similar each sample is to each other sample.
Secondly, each sample sends an availability message to the other samples reflecting how well suited said sample is to act as exemplar for them.
Each message type influences the message sent by the other iteratively until the system stabilises or reaches an end point with a set number of representative exemplars each with a set of samples clustered as being similar to them.
For full details regarding the implementation of this algorithm, readers are directed to the original work~\cite{Frey2006}.
This method is not randomised and so gives consistent results.
It does not require a number of clusters to be defined in advance.
These are both desirable traits within our work as we desire consistency and cannot make assumptions about the number of distinct groups which should be created.
The only parameters which we define in advance is the similarity matrix used for initialisation and a dampening factor which effects how heavily each new iteration updates the previous iteration's responsibility and availability values.
In our work, we use a cosine similarity matrix and a dampening factor of 0.5.
Other dampening factors were tested however these had little effect on the achieved results.

\vskip 3mm

\noindent \textbf{Locked Hierarchical Clustering}

\noindent Another method for clustering used within this work is locked hierarchical clustering.
This is the variation to agglomerative hierarchical clustering~\cite{Johnson1967} that was introduced in a previous work~\cite{Hall2017}.
This method, like AP is able to operate without a predefined number of clusters, instead slowly grouping the samples/clusters which are closest in their descriptor space together until a set ending criterion is reached.
We use symmetric Kullback-Leibler (KL2) distance~\cite{Siegler1997} to determine which samples/clusters are closest, treating each cluster as a multivariate Gaussian in order to make the calculation.
The delta Bayesian information criterion ($\Delta$BIC)~\cite{Schwarz1978} is what we used as the ending criterion, dictating that we stop merging if $\Delta$BIC becomes negative if the two closest clusters were to be merged.
Instead of initialising every plant image as being separate, the locking criterion used in this variation of agglomerative clustering ensures that multiple images of a sample plant are grouped together at initialisation.
In instances where this is not possible, this algorithm is initialised with every sample being separate as is normal for agglomerative hierarchical clustering~\cite{Johnson1967}.
More details regarding this algorithm can be found in~\cite{Hall2017}.
As KL2 distance and $\Delta$BIC are metrics based upon the mean and standard deviations of multivariate Gaussian representations of clusters, we predefine the mean and standard deviation of any isolated, unclustered sample.
For said samples we set the mean as the location of the sample in the descriptor space and the standard deviation as the median Euclidean distance between all samples given to the clustering algorithm.

\vskip 3mm

\noindent \textbf{K-Means Clustering}

\noindent K-means is a commonly used clustering algorithm where the number of clusters is predefined in advance~\cite{MacQueen1967}.
In our work it is used as a baseline method for showing the relation to the number of clusters/exemplars shown to the user and the resultant labelling/classification accuracy.
After a randomly initialising the cluster centres (cluster means), k-means iteratively refines the location of the mean of the cluster as samples iteratively change cluster allocation and update the mean of said clusters.
It is not considered as a methodology which would be ideal to use in the field as it requires a predefined number of clusters to be defined which is not realistic to know in advance.
The other disadvantage to using this technique is the random initialisation therein which cannot guarantee a consistent clustering result.
In our test, we perform tenfold clustering to minimise the effect of the random initialisation used in k-means and take the mean result for any algorithm evaluation results attained from k-means clustering.
We use the k-means++ algorithm to improve initialisation by ensuring a well spaced out initialisation stage~\cite{Arthur2007}.

Within this work, all of the aforementioned clustering algorithms are utilised in some combination with different labelling strategies to enable us to analyse and determine which approaches are best suited towards our task of an adaptable precision weeding strategy.



\subsubsection{Selective Data Labelling}\label{sec:method:label}
Selective data labelling is a key stage of the pipeline which enables farmers to utilise the information gathered in the field surveillance stage.
It is what enables us to rapidly deploy our classification system without need for prior assumptions about the weed species in the field.
The aim in this stage is to use as few samples as possible to correctly label the data collected by the robot, so as to diminish the work which has to be done by the farmer whilst maximising labelling accuracy as much as possible.
This is achieved through providing the user with exemplar representatives and utilising either clustering methods described previously or other approaches such as label propagation to label the remaining samples.
This is also the stage where a farmer would dictate which precision weeding approach should be utilised for each plant.

\begin{figure}[t]
    \centering
    \includegraphics[width=0.9\textwidth]{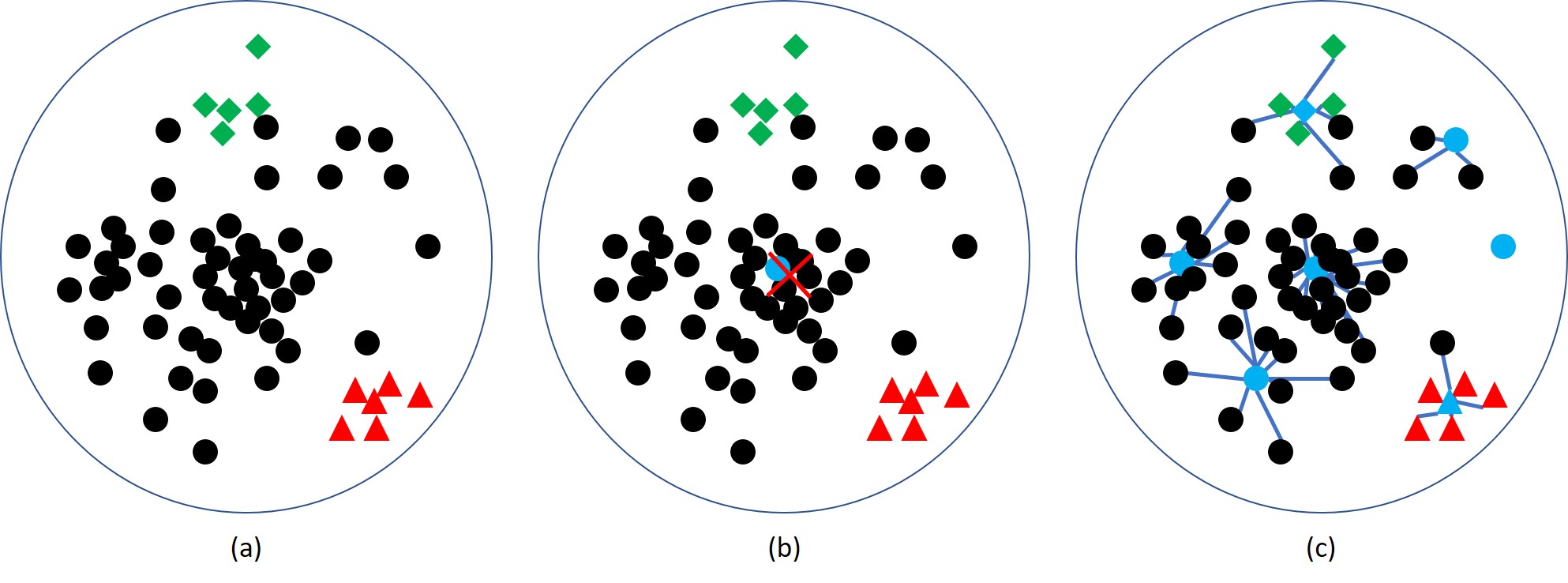}
    \caption{A simplified visual representation of two methods of exemplar generation. 
            (a) shows a single clustering result containing three different classes (black dots, red triangles and green diamonds). (b) shows mean exemplar generation where the exemplar (shown in blue) is selected as the sample closest to the mean (red cross). (c) represents exemplars calculated for AP-Refine (blue) as calculated through the AP algorithm with samples grouped with said exemplars represented by the lines joining samples to exemplars. Best viewed in colour.}
    \label{fig:method:labelling:exemplars}
\end{figure}

We examine several different approaches to labelling data including mean labelling, AP-refinement~\cite{Hall2017a}, and label propagation~\cite{Zhou2004}.
An example of mean labelling and AP-Refinement is shown in Figure~\ref{fig:method:labelling:exemplars}.

\vskip 3mm

\noindent \textbf{Mean Labelling}

\noindent Mean labelling is perhaps the simplest and most common of the methods used here.
After some initial clustering algorithm has sorted the data into clusters, a single exemplar sample for each cluster is presented, defined as the sample closest to the mean of the data in the descriptor space.
The label provided by the user for this exemplar is then defined as the label for each sample within the given cluster.
This is shown in a simplified way in Figure~\ref{fig:method:labelling:exemplars}-(b).
Mean labelling can be expanded upon to choose an arbitrary number of samples closest to the mean to give the user a stronger indication of the average representation of the cluster~\cite{Hall2017a} but that labelling method is not considered in this work.

\vskip 3mm

\noindent \textbf{AP-Refinement}

\noindent The labelling technique of AP-Refinement is a technique designed to show the variety of data contained within a given cluster rather than a centralised mean representation.
In this process, AP clustering\cite{Frey2006} is performed within each cluster to find a number of subclusters therein.
Each AP subcluster provides its own exemplar as part of the AP clustering process and these exemplars are shown to the user for labelling.
This refines upon the more broad clustering done initially.
This technique is shown to be particularly useful for dealing with impure clusters and can aid in identifying small subclusters of classes that do not fit the dominant class of the cluster.
A visual representation of how these AP subgroups could look in a given cluster is shown in Figure~\ref{fig:method:labelling:exemplars}-(c).
In our work, we calculate a separate similarity matrix for each cluster as the starting point for generating the appropriate exemplars for each cluster.
We use the same type of similarity matrix and dampening factor here as was described in Section~\ref{sec:method:clust}.

\vskip 3mm

\noindent \textbf{Label Propagation}

\noindent Label propagation, is a procedure described by Zhou et al.~\cite{Zhou2004}.
The procedure in effect iteratively updates label confidences for each sample based on how similar neighbouring samples are to them and these neighbour's own label confidences.
The similarity is based from an initial similarity graph and the label confidences from a set of known labels.
In our work we use a fully connected similarity graph using a Gaussian kernel with a standard deviation of 0.16 to describe similarities.
This similarity graph is then converted to a symmetrically normalised Laplacian graph which becomes the new similarity graph for the remainder of the algorithm.
The algorithm also contains a clamping factor which dictates the importance that each new iteration's label probability calculation should have on the final label probability.
This, in effect, adjusts the rate of convergence for the algorithm.
In our work we define this clamping factor as 0.2.
For further details we direct readers to the original work~\cite{Zhou2004}.

Due to the procedure for updating label confidences, sometimes the initially labelled samples can be relabelled by the algorithm, leading to incorrect data labelling.
This is a scenario that must be avoided within our work as we should never override the human decision. 
We therefore examine a variation of this algorithm where we lock the confidence of each manually labelled sample as 1, the highest confidence rating possible.
We shall refer to this procedure as locked label propagation.

Through different combinations of unsupervised weed clustering and selective labelling methods, this work attains insight into what methodologies are most appropriate for quickly labelling data attained during initial field surveillance.
The data collected and labelled using our techniques can then be used to train a classifier which will enable precision weeding for the final stage of our pipeline.

\subsubsection{Classifier Training}\label{sec:method:class}
The final stage in the offline data processing and labelling stage of our pipeline is using the labelled data to train a classification system.
This classification system, once trained, is what will be used in the final deployment of the precision weeding system.
In this work, we improve upon the classification accuracies presented in our previous work~\cite{Hall2017a} by utilising a DCNN classification system in this final stage.

Here, we adjust the output layer of our original bottleneck network described in Section~\ref{sec:method:feat} to match the classes provided by the user during the labelling stage of the system.
Using the connection weights attained from the initial fine-tuning for feature extraction using the PlantCLEF data as a baseline, we further fine-tune the network on the new data collected in the field.

To increase robustness, we perform data augmentation on the labelled training data.
We apply rotation, flipping, cropping and illumination variation to our data.
For rotation, we considered a rotation angle of $\theta = \{5, 10, 90, 180, 270, 350, 355\}\degree$ to show the system a combination of slight variation to angle as well as the larger orientation shifts of 90$\degree$.
Flipping was conducted over the $x$ and $y$ axes but only to the original image.
Cropping was done to augment the original data in a manner representative of a poor region proposal within plant detection.
We cropped the original image to 80\% and 90\% of the original image size.
Our final augmentation to our data was an adjustment to the illumination of plant pixels.
Here, we utilise the mask generated from plant segmentation to ensure that we only adjust the plant pixels and not the segmented black background.
Adjustment of illumination was achieved by transforming the image to the Hue-Saturation-Value (HSV) colour space and changing the Value of plant pixels to $V_{new}=\{0.8,0.9,1.1,1.2\}\times V{o}$ where $V{o}$ is the original value that each plant pixel had.
In combination with the original image collected from the field, this provided 16 training images for each image in the original training set.

Once the DCNN has been trained on the augmented labelled data in the offline processing stage of our overall pipeline, it is ready to be implemented as a part of the final precision weeding system.

\subsection{Automated Precision Weeding}
\begin{figure}[t]
    \centering
    \includegraphics[width=0.95\textwidth]{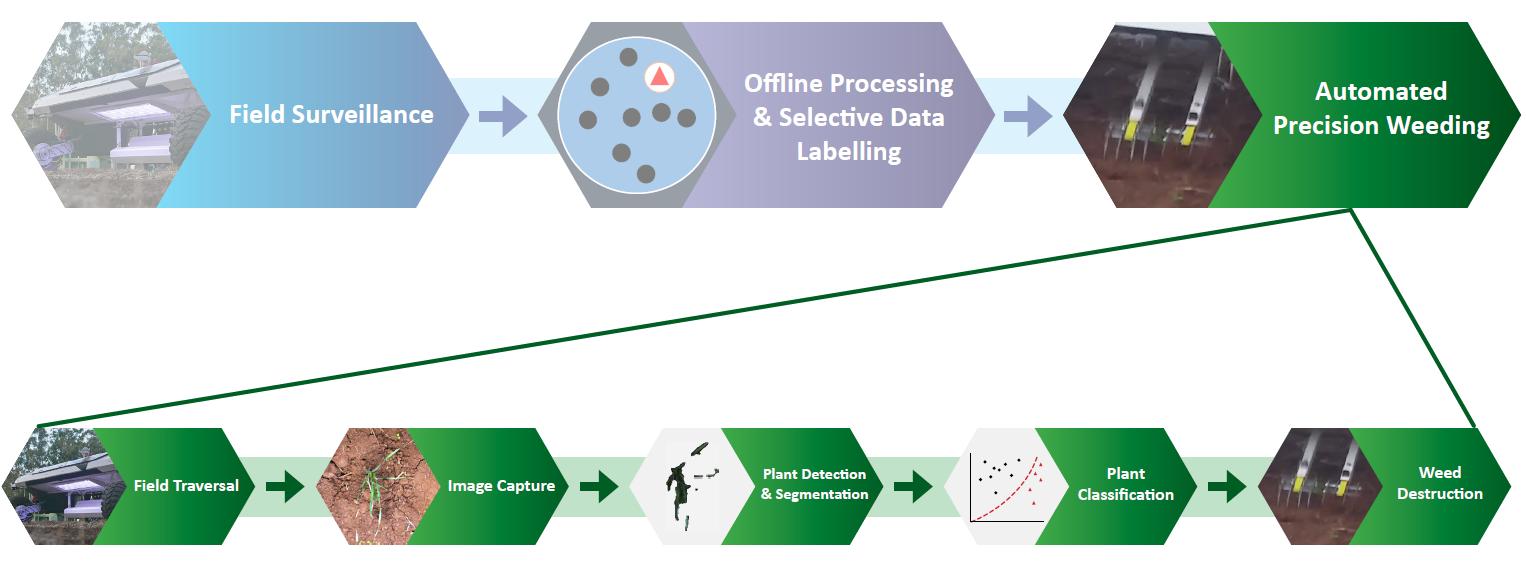}
    \caption{Automated precision weeding pipeline of stage three of main pipeline}
    \label{fig:method:pipeline:weeding}
\end{figure}

The final stage of the pipeline is the final deployed automated precision weeding process summarised in Figure~\ref{fig:method:pipeline:weeding}.
This process consists of detecting and segmenting plant regions, classifying identified plants and performing the appropriate weed destruction technique for the species identified for each plant.

The process within this final stage utilises procedures described in previous sections.
As mentioned previously, we do not focus on the field traversal and image capture stage as these are not within the scope of this work.
The detection and segmentation taken place is identical to that used within the initial field surveillance stage described in more detail in Section~\ref{sec:method:detect}.
The classification is performed using the algorithm which was previously trained within the offline processing and data labelling stage described in Section~\ref{sec:method:class}.
The following sections shall briefly describe the final deployment of our weed classification system and the weed destruction stage respectively.

\subsubsection{Plant Classification}\label{sec:method:class_deploy}
In this final stage of the overall pipeline, the trained DCNN described in Section~\ref{sec:method:class} is fed individual segmented plant regions and provides a score for each class which has been defined by the user based upon the output of the final layer.
The class with the highest score is then stated as the class of the plant.
In cases where there are multiple images available for a plant, we calculate the sum of the scores for each image.
Final class is the class with the highest summed score.

This classification system is what could finally be implemented on an agricultural robotics platform to perform automated, species-specific precision weeding based upon farmer defined weeding methods for each species.

\subsubsection{Weed Destruction}\label{sec:method:destruct}
The final stage in our proposed pipeline is the weed destruction stage where a method of weed management deemed most appropriate for each given weed species is applied when said species is identified using our weed classification.
The efficacy of the weeding approaches which can be used in this stage vary depending on a farmer's resources and desired management approach.
Because of these variances, we do not evaluate weed management efficacy for this stage in this work.
Weeding efficacy of different weed management approaches is its own challenging field of study and is beyond the scope of our work.
Instead we evaluate the classification accuracy of the system which would be deployed in this final stage.
This accuracy demonstrates the effectiveness of our system in successfully classifying plant species without any prior knowledge of what those species were at the beginning of our pipeline.
It is this classification accuracy which we shall use as an approximation to demonstrate how effective the weed destruction would be if implemented in the final stage but without needing to consider the variables that go into the decision for final weed management strategy for each species.

\section{Experimental Setup}\label{sec:setup}
In this work we test and evaluate several selective labelling methodologies and see how they affect the final classification accuracy of our system.
Using field data we step through our entire weed classification within our proposed pipeline.
We focus on the classification stage of the pipeline rather than the final deployed precision weed management as the effectiveness of such weed management can vary depending on a farmer's desired approach to precision weed management and the resources available to them.
The critical stage that our pipeline would need to accomplish successfully in order for a farmer to implement their desired weed management strategy is accurate species-wise weed classification.
The system will need to show itself to be able to operate rapidly with few selected labels being shown to the user, and without any prior knowledge or assumptions about the plant species that the system will encounter before initial scouting.
The following subsections shall describe the datasets, selective labelling methods, and evaluation metrics used to analyse our classification system.

\subsection{Datasets}\label{sec:setup:data}
Our work was performed on two separate plant datasets.
These shall be referred to as the Redlands and Flavia datasets.

\begin{figure}[t]
	\centering
	\setlength{\fboxsep}{0pt}
	
	\begin{subfigure}[b]{0.2\linewidth}
		\centering
		\fbox{\includegraphics[width=\linewidth, height=3cm]{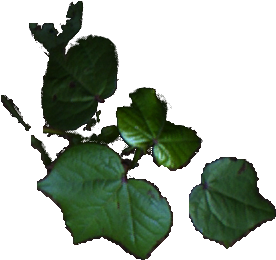}}
		\caption{Cotton}
	\end{subfigure}
	\begin{subfigure}[b]{0.2\linewidth}
		\centering
		\fbox{\includegraphics[width=\linewidth, height=3cm]{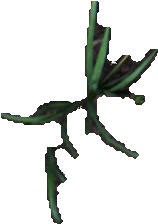}}
		\caption{Feathertop}
	\end{subfigure}
	\begin{subfigure}[b]{0.2\linewidth}
		\centering
		\fbox{\includegraphics[width=\linewidth, height=3cm]{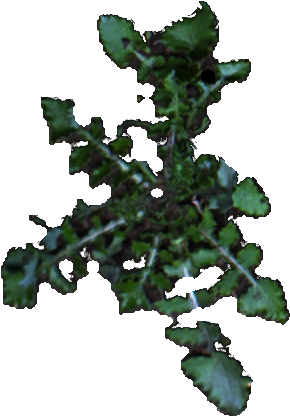}}
		\caption{Sowthistle}
	\end{subfigure}
	\begin{subfigure}[b]{0.2\linewidth}
		\centering
		\fbox{\includegraphics[width=\linewidth, height=3cm]{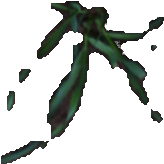}}
		\caption{Wild Oats}
	\end{subfigure}
	\caption{Example of each plant species within Redlands dataset. Black segmented backgrounds have been replaced with white and images have undergone slight distortion for visualisation purposes. Best viewed in colour.}
	\label{fig:materials:redlands_imgs}
\end{figure}

\begin{figure}
    \centering
    \includegraphics[width=0.9\linewidth]{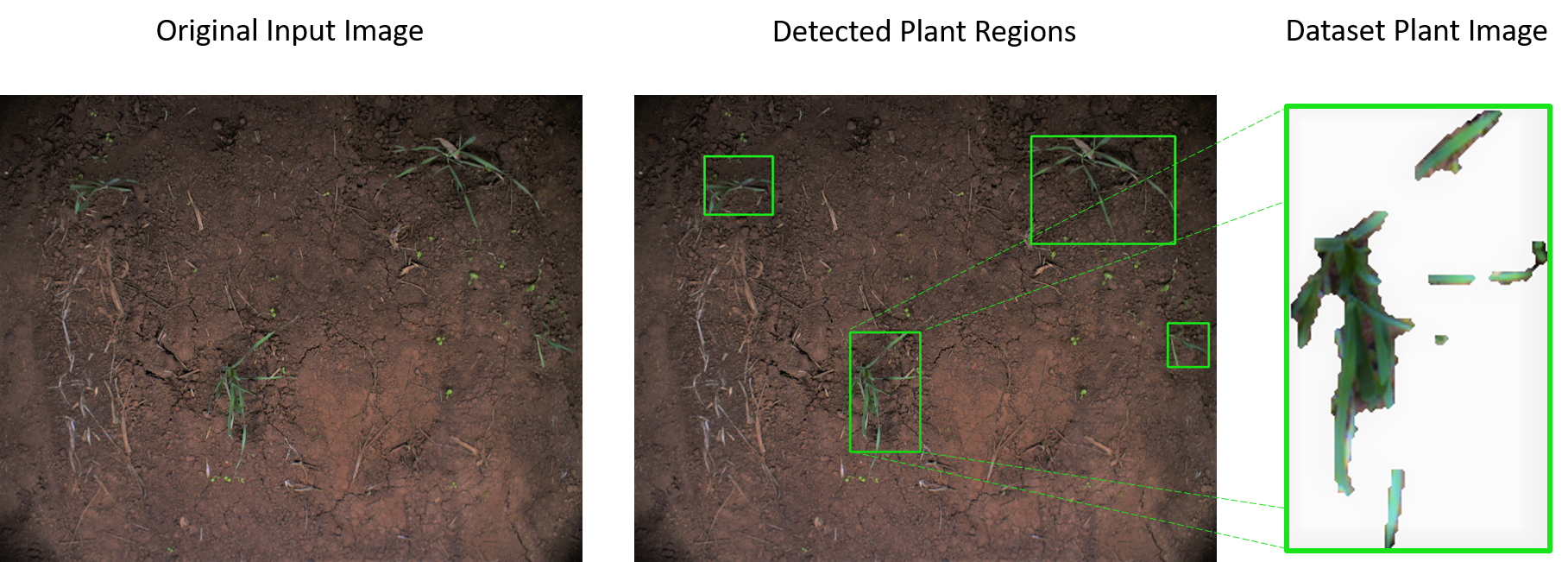}
    \caption{Example of dataset generation for the Redlands dataset. Best viewed in colour}
    \label{fig:setup:red_generation}
\end{figure}

The Redlands dataset consists of images collected using the AgBot II agricultural robotic platform~\cite{Bawden2017} on a small field.
Data was collected using a downwards facing IDS UI-1240SE 1.3MP global shutter camera rig under the AgBot II, with a field of view illuminated using a pulsed lighting system synchronised with the data capture.
This collected colour images at a rate of 5 Hz as the robot traversed a field populated with four different weed species.
The weed species observed were cotton (genus Gossypium), feathertop (Chloris virgata), sowthistle (Sonchus oleraceus) and wild oats (Avena fatua), examples of which can be viewed in Fig~\ref{fig:materials:redlands_imgs}.
Plants were detected and segmented using the approach described in Section~\ref{sec:method:detect} from the original images to generate all images in the dataset such as is shown in Figure~\ref{fig:setup:red_generation}.
As the robot traversed the field, multiple images of each plant were often taken.
We manually made note of which images were associated with individual plants for use in our experiments, simulating the effect of plant tracking and localisation which can be achieved using such techniques as precise GPS positioning of each plant.
Data was collected on two separate days three weeks apart and the data from both days was combined to form one dataset.
This combined dataset was then split into two parts, a training set and a testing set.
These are used to simulate a scenario where a robot is initially sent out to scout an area (training) and is then used to detect and treat weeds after the system has been trained (testing) associated with the first and last stages of our main pipeline outlined in Section~\ref{sec:method}.
Each contains 50\% of the individual plants observed for each species.
This is the same data utilised within~\cite{Hall2017a} and is summarised in Table~\ref{tbl:materials:redlands}.

\begin{table}[t]
	\centering
	\caption{Redlands dataset summary.}
	\label{tbl:materials:redlands}
	\begin{tabular}{|c|c|c|}
		\hline
		& \textbf{Training} & \textbf{Testing} \\ \hline
		\textbf{Cotton}     & 134            & 132            \\ \hline
		\textbf{Feathertop} & 128             & 126            \\ \hline
		\textbf{Sowthistle} & 25             & 28             \\ \hline
		\textbf{Wild Oats}  & 131            & 126            \\ \hline
	\end{tabular}
\end{table}

The Flavia dataset was originally published in~\cite{Wu2007} and consists of 1907 scanned leaf images on white backgrounds covering 32 different plant species.
Although this presents a more constrained view for an agricultural robotics setting when compared to the Redlands dataset, the Flavia dataset provides many more images and a greater variety of distinct species and is a commonly used dataset in leaf classification literature~\cite{Wang2017}~\cite{Lee2017}~\cite{Hall2015}.
For the segmentation stage of our pipeline, we use a different segmentation method than that presented in Section~\ref{sec:method:detect} and instead use the procedure outlined in the original paper~\cite{Wu2007}.
This consists of applying a predefined threshold to a specific grayscale representation of each image.
This is used to provide the most ideal segmentation possible.
We also divide this dataset into a training and testing set to simulate an unsupervised initial scouting of plants followed by a final classification stage using the data from initial scouting.
We separate approximately 50\% of the images for each species into the training and testing sets.
This provided us with 977 training, and 930 testing images.
Because of the constrained environment these images were taken in, there is not multiple images of a single plant as was the case in the Redlands dataset.
A visual comparison of the two datasets is shown in Figure~\ref{fig:discussion:data_compare}.
Through use of both datasets, we tested and evaluated different offline processing and selective labelling techniques to be used within our pipeline to achieve high classification accuracy when the system is deployed for precision weed management.

\subsection{Evaluated Selective Labelling Methods}\label{sec:setup:label}
The data labelling techniques which can be used in our proposed pipeline and are outlined fully in Section~\ref{sec:method:label}, are closely linked to the clustering algorithm used prior to them.
In this work we expand upon our previous study~\cite{Hall2017a} which only considered one initial clustering procedure by examining several different combinations.
A summary of the approaches used and the test names which they shall be referred to as throughout the remainder of this work is given in Table~\ref{tbl:method:test_names}.
For simplicity, we assume that the human will correctly label any image presented to them.

Most of the labelling techniques described here are covered in more detail in Section~\ref{sec:method:label} with two exceptions.
Firstly there is the technique used in \AP, where the exemplars attained after performing AP clustering are shown to the user the label given to said exemplars are applied to their respective clusters.
Secondly, the \Full~experiment labels every single sample in the training set is labelled manually rather than using some selective labelling technique.
This is the baseline that shall provide the best possible labelling accuracy but which is undesirable for an applied system as it is the most time consuming and tedious for a human to use of all the methods presented.

For the experiments \LP~and \LLP, there is no clustering which is initially performed.
Instead, a set number of images are randomly sampled from the entire set of images to act as exemplars.
To account for the randomness in this experiment, we perform each test involving these methods with 10 separate random samplings and report upon the mean results of these tests.
The \APLP~and \APLLP~experiments both use the exemplars attained through affinity propagation as the samples to be labelled followed by label propagation to label all of the samples rather than trusting the clustering attained through AP as is done in the \AP~experiment

\begin{table}[t]
\centering
\caption{Clustering and labelling method test summary. Each test comprises of a clustering operation followed b a given labelling method. Clustering methods used are k-means clustering, locked hierarchical clustering, and affinity propagation (AP) clustering. Labelling methods include full labelling, mean exemplar-based labelling (Mean), affinity propagation refinement (AP-Refine) labelling, use of affinity propagation exemplars (AP-Exemplars), label propagation (LP), and locked label propagation (LLP).}\label{tbl:method:test_names}
\begin{tabular}{|c|c|c|}
\hline
\textbf{Test Name} & \textbf{Clustering Method} & \textbf{Labelling Method} \\ \hline
\textbf{Full}      & None                       & Full                      \\ \hline
\textbf{KMeans}    & k-means                    & Mean                      \\ \hline
\textbf{Mean}      & Locked Hierarchical        & Mean                      \\ \hline
\textbf{AP-Refine} & Locked Hierarchical        & AP-Refine                 \\ \hline
\textbf{AP}        & AP                         & AP-Exemplars              \\ \hline
\textbf{LP}        & None                       & Label Propagation         \\ \hline
\textbf{LLP}       & None                       & Locked Label Propagation  \\ \hline
\textbf{APLP}      & AP                         & Label Propagation         \\ \hline
\textbf{APLLP}     & AP                         & Locked Label Propagation  \\ \hline
\end{tabular}
\end{table}

In the case where there are multiple images of a single plant such as in the Redlands dataset, some of the techniques will not provide the same label to each image of said plant.
When using algorithms that meet this criteria, we use a majority vote system where the class associated with the majority of the images of a given plant is considered as the class for all images of the plant.

\subsubsection{Choosing Number of Exemplars}
One of the focuses in our experiments is on achieving accurate labelling whilst selecting a low number of samples for labelling.
For \Mean, \APR, \AP, \APLP, and \APLLP~the number of samples is calculated automatically as a part of their clustering algorithms.
The \KM, \LP, and \LLP~however, all require a predefined number of clusters or exemplars.
We present results for these methods using a range of different number of exemplars.
This gives us a reference for the effect that the number of samples selected for labelling has on the labelling and classification accuracy in general.
For \LP~and \LLP~we also demonstrate a scenario where the number of exemplars is equal to the number of exemplars attained through \AP~and \APR.
This is to allow a more direct comparison between a randomised set of exemplars and a more structured choice of exemplar.
As the fine-tuning process that we use for training our classifier can be quite time-intensive, we only calculate the lowest numbers of labels (5\%, 10\%, and 20\%) for \KM~for the classification stage.
We also calculate \LP~and \LLP~ for labelling 10\% and 20\% of the data as well as the scenarios where the number of exemplars is equal to that of \AP~and \APR.
Our analysis will focus most heavily on \LP~and \LLP~results with an equal number of clusters to \AP~as these are directly comparable to the results attained using \APLP~and \APLLP.
We also demonstrate classification accuracy for \Full~to show the maximum classification accuracy that we could expect when using the fully labelled training set.

\subsection{Evaluation Metrics}
In order to test the efficacy of our proposed pipeline and the selective labelling techniques used therein, we evaluate in terms of labelling accuracy and classification accuracy across both the Redlands and Flavia datasets.
As our focus is on trying to label and classify plants with as little human interaction as possible, we summarise labelling and classification accuracy versus the number of exemplars labelled by humans.

Labelling accuracy is the rate at which we correctly label samples from our initial scouting data.
This is the critical stage of the pipeline as the labelling accuracy will heavily influence the final classification accuracy.
Labelling accuracy is tested separately on training data for the Redlands and Flavia datasets accordingly.
The maximum labelling accuracy that can be achieved is 100\% accuracy signifying that every image of every plant observed in the initial field surveillance stage was given the correct species label.

The other evaluation of note which we perform is classification accuracy of the deployed classification system.
Even if our pipeline is able to achieve 100\% labelling accuracy, the classification accuracy is not guaranteed to be 100\% due to the nature of classification algorithms and variation between training and testing data.
Classification accuracy quantifies how well our system, trained on data labelled during the offline processing and selective data labelling stage of our pipeline, is able to classify plants which it encounters in the field upon deployment.
Here, we currently assume that the plants found at this later stage are the same species as the plants found at initial scouting.
The classification accuracy would be the factor which most heavily dictates the weeding efficiency and efficacy which could be achieved within our proposed pipeline as it is what would be used by the automated weeding system to decide which weeding strategy should be used for a given plant.
As our work is focused on the rapidly deployable classification system which can enable a precision weeding strategy and not the effectiveness of different weeding methods on specific species, we do not test weeding efficacy.
Works showing real-time weeding effectiveness on the AgBotII platform are available in~\cite{Bawden2017}.
Classification accuracy is tested on the testing sets for each dataset, using classifiers trained using the labels generated by the labelling techniques defined previously.
It should be reiterated that before initial scouting, the classification system being tested had no knowledge of the weeds with which we evaluate it on.
\section{Results and Discussion}\label{sec:discussion}

\subsection{Labelling Accuracy}\label{sec:dicsussion:label}
%

\begin{table}[t]
\centering
\caption{Main Redlands labelling accuracy results for all tested labelling scenarios. \textbf{Full} labels all of the training data manually, having the user examine every sample. \textbf{KMeans} performs k-means clustering and clusters are labelled using the mean samples of each. 
\textbf{AP} performs affinity propagation (AP) clustering and each cluster is labelled based upon the label of the AP exemplar attained for that cluster. Both \textbf{Mean} and \textbf{AP-Refine} perform locked hierarchical clustering before labelling.\textbf{Mean} uses the same labelling approach as \textbf{KMeans} after clustering and \textbf{AP-Refine} refines the labelling from the initial clustering using AP exemplars. \textbf{LP}, \textbf{LLP}, \textbf{APLP}, and \textbf{APLLP} all use label propagation (LP) or locked label propagation (LLP) for labelling, beginning with either random samples (\textbf{LP} and \textbf{LLP}) or AP exemplars (\textbf{APLP} and \textbf{APLLP}).}

\label{tbl:results:red_label}
\begin{tabular}{|c|c|c|}
\hline
\textbf{Test Name} & \textbf{\begin{tabular}[c]{@{}c@{}}\% Training Data \\ Labelled\end{tabular}} & \textbf{\begin{tabular}[c]{@{}c@{}}Labelling \\ Accuracy\\ (\%)\end{tabular}} \\  \hline
\textbf{KMeans}    & 10.0                                                                          & 70                                                                             \\ \hline
\textbf{KMeans}    & 5.0                                                                           & 67                                                                            \\ \hline
\textbf{Mean}      & 0.7                                                                           & 63                                                                            \\ \hline
\textbf{Full}      & 100                                                                           & 100                                                                           \\ \hline
\textbf{AP-Refine} & 13.9                                                                          & 75                                                                            \\ \hline
\textbf{LP}        & 8.1                                                                           & 71                                                                            \\ \hline
\textbf{LLP}       & 8.1                                                                           & 71                                                                            \\ \hline
\textbf{APLP}      & 8.1                                                                           & 79                                                                            \\ \hline
\textbf{APLLP}     & 8.1                                                                           & 79                                                                            \\ \hline
\textbf{AP}        & 8.1                                                                           & 74  \\ \hline                                                                           
\end{tabular}
\end{table}

\begin{table}[t]
\centering
\caption{Main Flavia labelling accuracy results for all tested labelling scenarios. \textbf{Full} labels all of the training data manually, having the user examine every sample. \textbf{KMeans} performs k-means clustering and clusters are labelled using the mean samples of each. 
\textbf{AP} performs affinity propagation (AP) clustering and each cluster is labelled based upon the label of the AP exemplar attained for that cluster. Both \textbf{Mean} and \textbf{AP-Refine} perform locked hierarchical clustering before labelling.\textbf{Mean} uses the same labelling approach as \textbf{KMeans} after clustering and \textbf{AP-Refine} refines the labelling from the initial clustering using AP exemplars. \textbf{LP}, \textbf{LLP}, \textbf{APLP}, and \textbf{APLLP} all use label propagation (LP) or locked label propagation (LLP) for labelling, beginning with either random samples (\textbf{LP} and \textbf{LLP}) or AP exemplars (\textbf{APLP} and \textbf{APLLP}).}
\label{tbl:results:flavia_label}
\begin{tabular}{|c|c|c|}
\hline
\textbf{Test Name} & \textbf{\begin{tabular}[c]{@{}c@{}}\% Training Data \\ Labelled\end{tabular}} & \textbf{\begin{tabular}[c]{@{}c@{}}Labelling \\Accuracy\\ (\%)\end{tabular}} \\ \hline
\textbf{KMeans}    & 10.0                                                                          & 96                                                                            \\ \hline
\textbf{KMeans}    & 5.0                                                                           & 93                                                                            \\ \hline
\textbf{Mean}      & 3.0                                                                           & 52                                                                            \\ \hline
\textbf{Full}      & 100                                                                           & 100                                                                           \\ \hline
\textbf{AP-Refine} & 15.4                                                                          & 97                                                                            \\ \hline
\textbf{LP}        & 4.3                                                                           & 65                                                                            \\ \hline
\textbf{LLP}       & 4.3                                                                           & 66                                                                            \\ \hline
\textbf{APLP}      & 4.3                                                                           & 93                                                                            \\ \hline
\textbf{APLLP}     & 4.3                                                                           & 93                                                                            \\ \hline
\textbf{AP}        & 4.3                                                                           & 95                                                                            \\ \hline
\end{tabular}
\end{table}

\begin{figure}
    \centering
    \fbox{\includegraphics[width=0.8\linewidth]{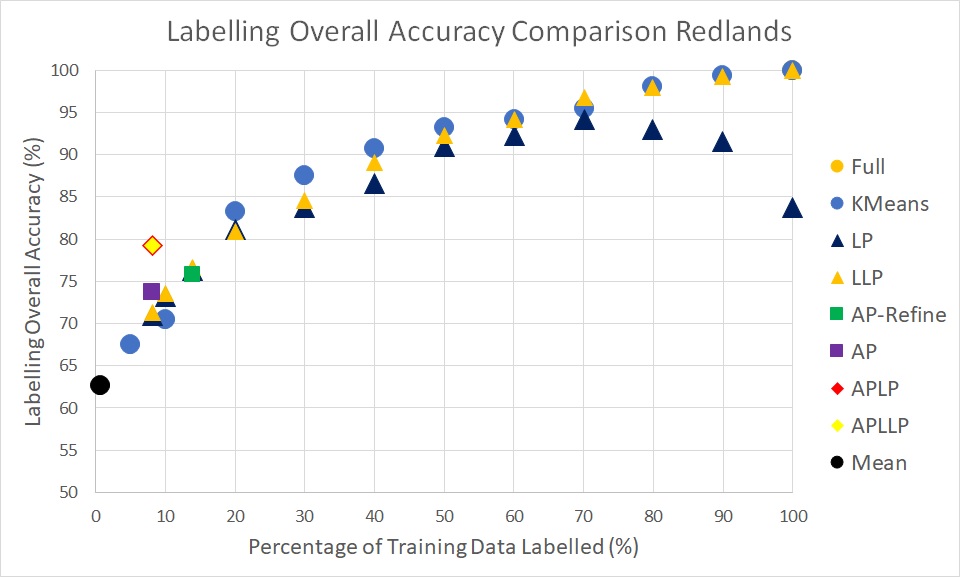}}
    \caption{Plot showing the overall labelling accuracy (y-axis) vs number of exemplars selected for labelling (x-axis) for the Redlands dataset across all tested scenarios including: full labelling (\textbf{Full}), k-means clustering followed by cluster mean labelling (\textbf{KMeans}), label propagation (LP) on randomly labelled samples (\textbf{LP}), locked label propagation (LLP) on randomly labelled samples (\textbf{LLP}), hierarchical clustering followed by affinity propagation (AP) refinement labelling (\textbf{AP-Refine}), AP clustering (\textbf{AP}), AP clustering followed by LP (\textbf{APLP}), AP clustering followed by LLP (\textbf{APLLP}), and locked hierarchical clustering followed by cluster mean labelling (\textbf{Mean}). Best viewed in colour.}
    \label{fig:results:red_label}
\end{figure}

\begin{figure}
    \centering
    \fbox{\includegraphics[width=0.8\linewidth]{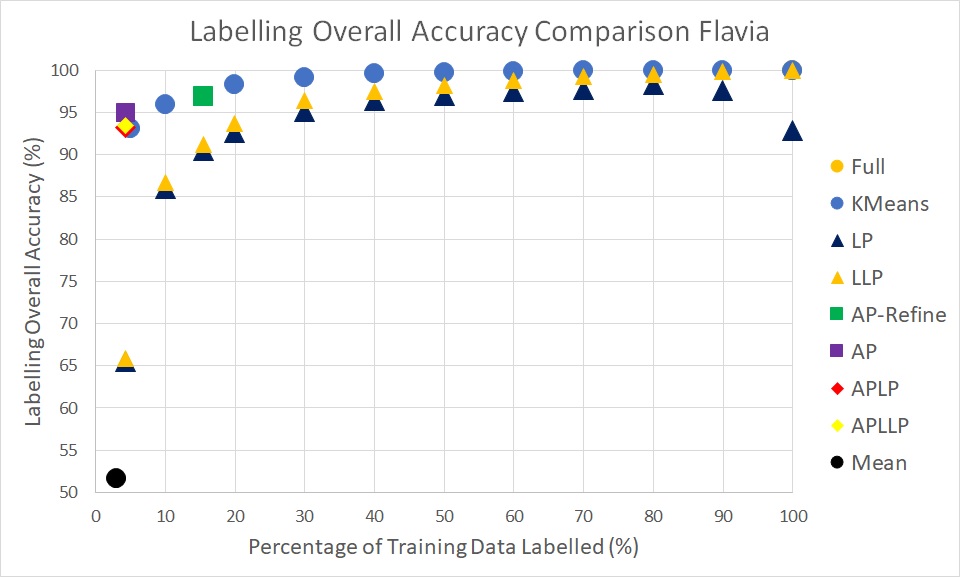}}
    \caption{Plot showing the overall labelling accuracy (y-axis) vs number of exemplars selected for labelling (x-axis) for the Flavia dataset across all tested scenarios including: full labelling (\textbf{Full}), k-means clustering followed by cluster mean labelling (\textbf{KMeans}), label propagation (LP) on randomly labelled samples (\textbf{LP}), locked label propagation (LLP) on randomly labelled samples (\textbf{LLP}), hierarchical clustering followed by affinity propagation (AP) refinement labelling (\textbf{AP-Refine}), AP clustering (\textbf{AP}), AP clustering followed by LP (\textbf{APLP}), AP clustering followed by LLP (\textbf{APLLP}), and locked hierarchical clustering followed by cluster mean labelling (\textbf{Mean}). Best viewed in colour.}
    \label{fig:results:flavia_label}
\end{figure}

The main labelling accuracy results we shall be analysing are presented in Tables~\ref{tbl:results:red_label} and~\ref{tbl:results:flavia_label} for the Redlands and Flavia datasets respectively.
The complete set of results attained is graphically shown for the Redlands and Flavia datasets in Figures~\ref{fig:results:red_label} and~\ref{fig:results:flavia_label} respectively.
Observing the results presented here we can make several observations regarding the effectiveness of different labelling algorithms within our proposed system.

When examining labelling accuracy across the graphical plots for both experiments, we see that \KM~labelling accuracy gradually increases as more exemplars are labelled.
The rate of change observed also seems to follow a curve.
Using this we also observe that \APR~seems to fit within this curve.
This suggests that \APR, even though it has a two-stage clustering procedure, produces similar labelling accuracy to what we would expect from a similar number of exemplars extracted through another clustering algorithm.
It should be noted however that unlike \KM, \APR~does not require the number of clusters to be defined in advance.

In both plots we can also observe the effect of locking the label propagation algorithm.
We see that both \LP~and \LLP~initially follow a very similar labelling accuracy curve.
However, as the number of labelled exemplars increases we see the labelling accuracy decrease for \LP~as apposed to \LLP.
This is because \LP, as described in Section~\ref{sec:method:label} allows the algorithm to overwrite labels given by the user.
As the number of exemplars increases, one would expect the labelling accuracy to also increase as more samples are manually allocated the correct label.
\LP~however has the ability to overwrite this correct label, oftentimes allowing the majority class to attain more samples than it rightly should.
For the Redlands dataset, we observed multiple instances where the sowthistle class, which has the lowest number of samples, was completely mislabelled even though multiple samples of sowthistle were selected for labelling.

We also notice across both datasets that the lowest performing system of all is the \Mean~system with a labelling accuracy of only 63\% and 52\% for Redlands and Flavia datasets respectively.
This can be attributed largely in part to the low number of exemplars being shown (as few as three exemplars for the  Redlands dataset).
It should however be noted that for the Redlands dataset, the result attained seems to follow the \KM~curve.
This suggests that the labelling results are actually fairly reasonable for the number of samples being shown to the user but without needing to define that number in advance of clustering.

Another observation that we can gather from the labelling results is that the only algorithms that were able to outperform the general trend shown by the \KM~curve were the results demonstrated by \AP, \APLP, and \APLLP.
These algorithms, particularly for the Redlands data, show a greater capacity for generating accurate labels with a low number of exemplars.
On the Redlands data collated in Table~\ref{tbl:results:red_label}, \AP, \APLP, and \APLLP~were able to achieve labelling accuracies of 74\%, 79\% and 79\% respectively with only 8.1\% of images selected for labelling.
However, we don't see this trend of exceeding the \KM~curve as significantly on the Flavia dataset.
Examining Table~\ref{tbl:results:flavia_label}, we see that \AP, \APLP, and \APLLP~do however achieve labelling accuracy equivalent to or exceeding \KM~when labelling 4.3\% of the available data whilst \KM~labels 5\% of the available data.
At these stages \KM~achieves 93\% labelling accuracy whilst \AP, \APLP, and \APLLP~achieve 95\%, 93\% and 93\% respectively.
Examining the labelling results in terms of number of exemplars labelled we conclude that the best performing systems for the Redlands data were \APLP, and \APLLP, which managed to attain 79\% labelling accuracy whilst labelling 12.3 times fewer images than full labelling.
In comparison to this we see that the best result for the Flavia dataset came from the \AP~labelling system which attained 95\% labelling accuracy whilst labelling 23.3 times fewer samples than full labelling.

\begin{figure}
    \centering
    \includegraphics[width=0.9\textwidth]{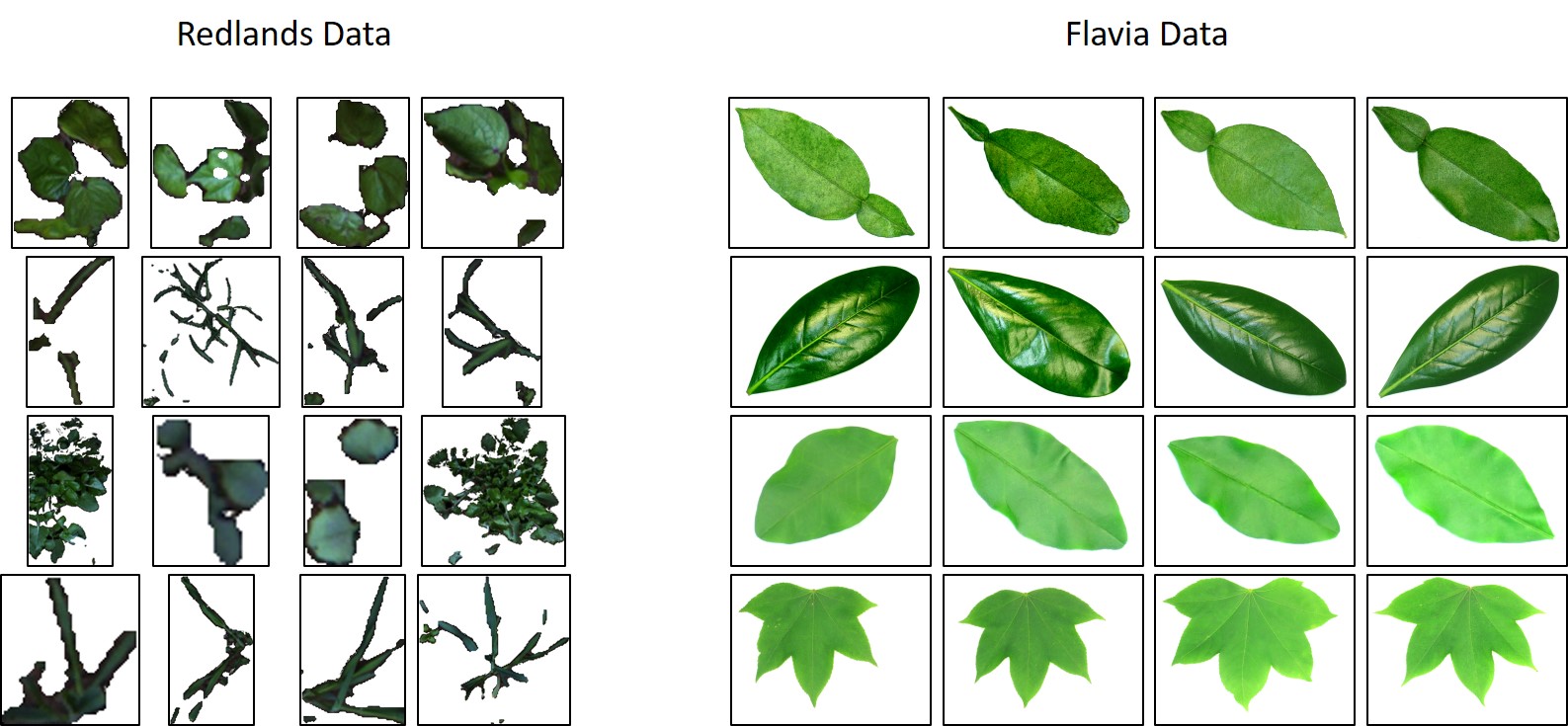}
    \caption{Comparison of interclass and intraclass variation within the Redlands and Flavia datasets. Interclass variation is seen as you change rows in either dataset and intraclass variation is seen as you change column. Backgrounds have been whitened and images rescaled and rotated for visualisation purposes. Best viewed in colour.}
    \label{fig:discussion:data_compare}
\end{figure}

Several times in our analysis, we observe that the results from the Flavia dataset typically outperform the results from the Redlands dataset.
We can also observe in Figure~\ref{fig:results:flavia_label} that in general the \KM~labelling accuracy does not drop as significantly for the Flavia dataset as was seen in the Redlands dataset.
We summize this is because of the cleaner image quality, more constrained environment and clearer distinction between many of the classes than in the Redlands data.
This can be shown visually in Figure~\ref{fig:discussion:data_compare} which demonstrates the intraclass variance and interclass similarity between the four classes in the Redlands training data when compared to four randomly chosen classes from the Flavia training data.
This difference between the datasets is our reasoning as to why AP-based labelling approaches were unable to exceed \KM~to the same extent which they had for the Redlands dataset and why there is a trend of higher labelling accuracy for the Flavia dataset test.
Focusing slightly more on the Redlands dataset with its more challenging field data, it is impressive that \AP, \APLP, and \APLLP~are able to achieve such a high labelling accuracy with so few images and exceeding the result of any of the other algorithms which sit within or below the k-means curve.

We also notice a great improvement for label propagation methods when using AP as a starting point for the algorithm.
Across both datasets, we see that the results for \LP~and \LLP~are surpassed by \APLP~and \APLLP~respectively when labelling the same amount of data.
This information is easily observed in Tables~\ref{tbl:results:red_label} and~\ref{tbl:results:flavia_label} where we see improvements of up to 28\% between the randomly initialised label propagation methods and those with AP performed for initialisation with the same number of samples.
This is due to the improved initialisation state that the algorithm is provided with through AP which gives some initial idea of the structure of data in the feature space before undergoing the automated label propagation procedure.
We see that when compared to \AP~however, that the \APLP~and \APLLP~algorithms only exceed \AP~for the Redlands dataset, correcting some of the labels which were misassigned by the actual AP algorithm.
This does not hold true for the Flavia dataset however, with all algorithms having very similar labelling accuracies of 95\%, 93\% and 93\% for \AP, \APLP, and \APLLP~respectively.

\subsection{Classification Accuracy}\label{sec:discussion:class}
%

\begin{table}[t]
\centering
\caption{Main Redlands classification accuracy results for all tested labelling scenarios. \textbf{Full} labels all of the training data manually, having the user examine every sample. \textbf{KMeans} performs k-means clustering and clusters are labelled using the mean samples of each. 
\textbf{AP} performs affinity propagation (AP) clustering and each cluster is labelled based upon the label of the AP exemplar attained for that cluster. Both \textbf{Mean} and \textbf{AP-Refine} perform locked hierarchical clustering before labelling.\textbf{Mean} uses the same labelling approach as \textbf{KMeans} after clustering and \textbf{AP-Refine} refines the labelling from the initial clustering using AP exemplars. \textbf{LP}, \textbf{LLP}, \textbf{APLP}, and \textbf{APLLP} all use label propagation (LP) or locked label propagation (LLP) for labelling, beginning with either random samples (\textbf{LP} and \textbf{LLP}) or AP exemplars (\textbf{APLP} and \textbf{APLLP}).}
\label{tbl:results:red_class}
\begin{tabular}{|c|c|c|}
\hline
\textbf{Test Name} & \textbf{\begin{tabular}[c]{@{}c@{}}\% Training Data \\ Labelled\end{tabular}} & \textbf{\begin{tabular}[c]{@{}c@{}}Classification \\ Accuracy\\ (\%)\end{tabular}} \\ \hline
\textbf{KMeans}    & 10.0                                                                          & 70                                                                                 \\ \hline
\textbf{KMeans}    & 5.0                                                                           & 69                                                                                 \\ \hline
\textbf{Mean}      & 0.7                                                                           & 68                                                                                 \\ \hline
\textbf{Full}      & 100                                                                           & 92                                                                                 \\ \hline
\textbf{AP-Refine} & 13.9                                                                          & 73                                                                                 \\ \hline
\textbf{LP}        & 8.1                                                                           & 68                                                                                 \\ \hline
\textbf{LLP}       & 8.1                                                                           & 68                                                                                 \\ \hline
\textbf{APLP}      & 8.1                                                                           & 78                                                                                 \\ \hline
\textbf{APLLP}     & 8.1                                                                           & 76                                                                                 \\ \hline
\textbf{AP}        & 8.1                                                                           & 77 \\  \hline                                                                               
\end{tabular}
\end{table}

\begin{table}[]
\centering
\caption{Main Flavia classification accuracy results for all tested labelling scenarios. \textbf{Full} labels all of the training data manually, having the user examine every sample. \textbf{KMeans} performs k-means clustering and clusters are labelled using the mean samples of each. 
\textbf{AP} performs affinity propagation (AP) clustering and each cluster is labelled based upon the label of the AP exemplar attained for that cluster. Both \textbf{Mean} and \textbf{AP-Refine} perform locked hierarchical clustering before labelling.\textbf{Mean} uses the same labelling approach as \textbf{KMeans} after clustering and \textbf{AP-Refine} refines the labelling from the initial clustering using AP exemplars. \textbf{LP}, \textbf{LLP}, \textbf{APLP}, and \textbf{APLLP} all use label propagation (LP) or locked label propagation (LLP) for labelling, beginning with either random samples (\textbf{LP} and \textbf{LLP}) or AP exemplars (\textbf{APLP} and \textbf{APLLP}).}
\label{tbl:results:flavia_class}
\begin{tabular}{|c|c|c|}
\hline
\textbf{Test Name} & \textbf{\begin{tabular}[c]{@{}c@{}}\% Training Data \\ Labelled\end{tabular}} & \textbf{\begin{tabular}[c]{@{}c@{}}Classification \\ Accuracy \\ (\%)\end{tabular}} \\ \hline
\textbf{KMeans}    & 10.0                                                                          & 97                                                                                 \\ \hline
\textbf{KMeans}    & 5.0                                                                           & 95                                                                                 \\ \hline
\textbf{Mean}      & 3.0                                                                           & 51                                                                                 \\ \hline
\textbf{Full}      & 100                                                                           & 99                                                                                 \\ \hline
\textbf{AP-Refine} & 15.4                                                                          & 98                                                                                 \\ \hline
\textbf{LP}        & 4.3                                                                           & 67                                                                                 \\ \hline
\textbf{LLP}       & 4.3                                                                           & 67                                                                                 \\ \hline
\textbf{APLP}      & 4.3                                                                           & 94                                                                                 \\ \hline
\textbf{APLLP}     & 4.3                                                                           & 95                                                                                 \\ \hline
\textbf{AP}        & 4.3                                                                           & 97               \\  \hline                                                                 
\end{tabular}
\end{table}

\begin{figure}
    \centering
    \fbox{\includegraphics[width=0.8\linewidth]{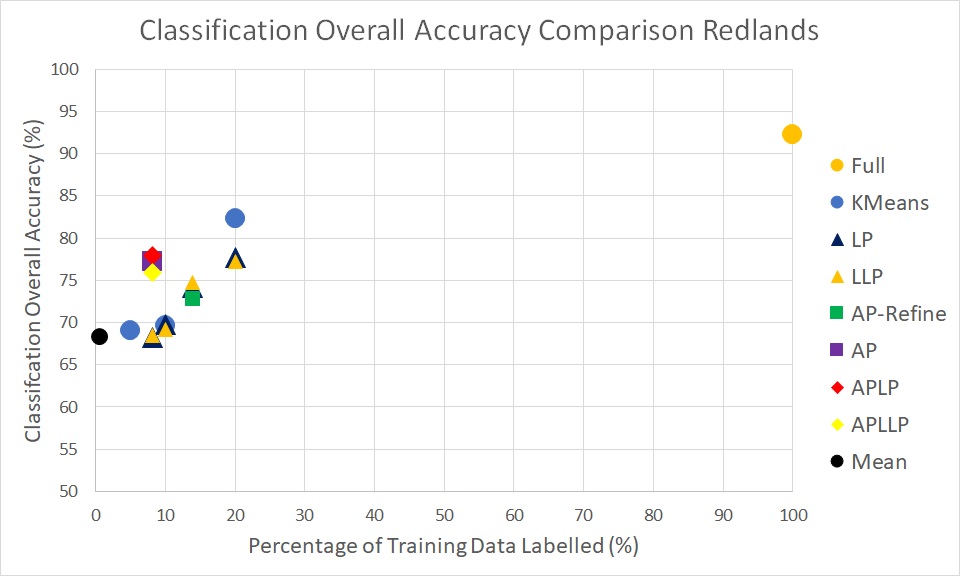}}
    \caption{Plot showing the overall classification accuracy (y-axis) vs number of exemplars selected for labelling (x-axis) for the Redlands dataset across tested scenarios including: full labelling (\textbf{Full}), k-means clustering followed by cluster mean labelling (\textbf{KMeans}), label propagation (LP) on randomly labelled samples (\textbf{LP}), locked label propagation (LLP) on randomly labelled samples (\textbf{LLP}), hierarchical clustering followed by affinity propagation (AP) refinement labelling (\textbf{AP-Refine}), AP clustering (\textbf{AP}), AP clustering followed by LP (\textbf{APLP}), AP clustering followed by LLP (\textbf{APLLP}), and locked hierarchical clustering followed by cluster mean labelling (\textbf{Mean}). Best viewed in colour.}
    \label{fig:results:red_class}
\end{figure}

\begin{figure}
    \centering
    \fbox{\includegraphics[width=0.8\linewidth]{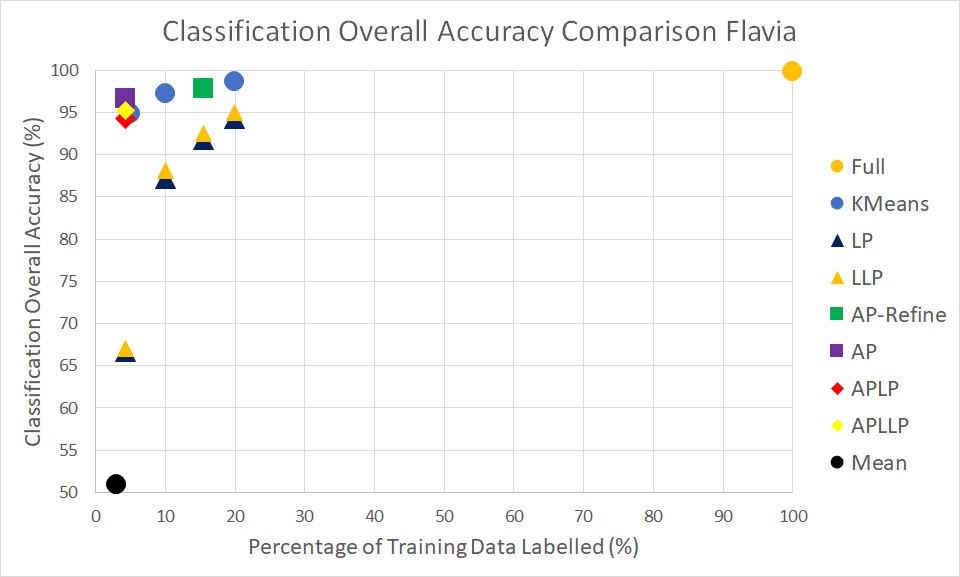}}
    \caption{Plot showing the overall classification accuracy (y-axis) vs number of exemplars selected for labelling (x-axis) for the Flavia dataset across tested scenarios including: full labelling (\textbf{Full}), k-means clustering followed by cluster mean labelling (\textbf{KMeans}), label propagation (LP) on randomly labelled samples (\textbf{LP}), locked label propagation (LLP) on randomly labelled samples (\textbf{LLP}), hierarchical clustering followed by affinity propagation (AP) refinement labelling (\textbf{AP-Refine}), AP clustering (\textbf{AP}), AP clustering followed by LP (\textbf{APLP}), AP clustering followed by LLP (\textbf{APLLP}), and locked hierarchical clustering followed by cluster mean labelling (\textbf{Mean}). Best viewed in colour.}
    \label{fig:results:flavia_class}
\end{figure}

Using classifiers trained on the data labelled data from the previous stage, we examine the classification accuracies for each method presented.
The main results are summarised in Tables~\ref{tbl:results:red_class} and~\ref{tbl:results:flavia_class} and a graphical view of the classification results attained is shown in Figures~\ref{fig:results:red_class} and~\ref{fig:results:flavia_class}.
We see here many of the trends we had previously identified for labelling accuracy with some exceptions.

In most instances, we see the classification accuracy very closely resembles the labelling accuracy but with two main variations.
These variations exist within the Redlands data which is reasonable considering the greater level of variance within images from said dataset.
We observe in Table~\ref{tbl:results:red_class} that \Full~labelling only attained 92\% accuracy even though it achieved 100\% labelling accuracy.
This is considered to occur due to the aforementioned variation between the training and testing datasets which is more prominent in the Redlands dataset than the Flavia dataset.
The other change of note within the Redlands tests comes from \Mean~labelling which has an increase of classification accuracy of 68\% from the original labelling accuracy of 63\%, placing it slightly above the expected trajectory of the \KM~curve.
We also notice a small change in \AP, \APLP, and \APLLP~results as they are now closer to equal in classification accuracy, achieving classification scores of  77\%, 78\%, and 76\% respectively.
This seems to demonstrate that while both \APLP~and \APLLP~achieved high labelling accuracy rates, that the choice of samples labelled correctly or incorrectly in these cases was actually of more detriment than those of \AP.
When examining the true positive rates of each class in the Redlands data for these experiments as is shown in Table~\ref{tbl:results:class_tpr_red_AP}, we notice that \APLP~and \APLLP~techniques seemed able to correctly classify cotton plants far better than \AP, but at a detriment to their feathertop and sowthistle accuracies.

\begin{table}[t]
\centering
\caption{True positive rate for affinity propagation (AP) clustering followed by locked label propagation (LLP) (\APLLP), AP clustering followed by label propagation (LP) (\APLP), and AP clustering labelled using AP exemplars (\AP)s~for the Redlands dataset}
\label{tbl:results:class_tpr_red_AP}
\begin{tabular}{|c|c|c|c|c|}
\hline
\multirow{2}{*}{\textbf{Test Name}} & \multicolumn{4}{|c|}{\textbf{True Positive Rates (\%)}} \\ \cline{2-5}
                                    & \textbf{Cotton} & \textbf{Feathertop} & \textbf{Sowthistle} & \textbf{Wild Oats} \\ \hline
\textbf{APLLP} & 92 & 78 & 75 & 58 \\ \hline
\textbf{APLP} & 95 & 75 & 86 & 61 \\ \hline
\textbf{AP} & 84 & 90 & 86 & 56 \\ \hline
\end{tabular}
\end{table}

Other than these small deviations, most of the trends and observations made from our labelling accuracy analysis apply to the classification tests, particularly for the Flavia dataset.
Comparing all classification accuracies to the labelling accuracies between Tables~\ref{tbl:results:red_class} and \ref{tbl:results:red_label}, and Tables~\ref{tbl:results:flavia_class} and \ref{tbl:results:flavia_label} for the Redlands and Flavia datasets respectively, we see an average difference of 2.7\% and 1.8\% respectively for each dataset.
This goes towards proving an important aspect of our approach.
None of the tests performed had any prior knowledge of the species in advance of our system being implemented.
The final accuracy we were able to achieve was in most cases close to identical to how well our data had been grouped, and by extension labelled, within the offline processing stage of our pipeline.
This is further evidence towards the idea that as long as an untrained system is able to distinguish different plant groups without needing any prior knowledge as to precisely what those plants are, we should be able to train the system to automatically recognise and treat those same species from then on.
It shows an approach such as ours could feasibly work  as a method for enabling a more rapidly deployable autonomous precision weeding systems capable of adapting to new fields and weeds as necessary without any prior weed species assumptions, attaining classification accuracies of 78\% and 97\% respectively for our field Redlands and Flavia leaf datasets.
This is a decrease of only 14\% and 3\% when compared to fully labelled data when selecting 12.3 and 23.3 times fewer images for labelling.

The main limitation to our approach is in the accuracy level attained in the offline processing and labelling stage of our pipeline.
While the results we attained are promising, more would be required before our system becomes an implementable solution.
Future work should focus on increasing the labelling accuracy that we can achieve, providing valuable exemplars in an unsupervised or semi-supervised manner so that autonomous precision weeding can operate without any prior knowledge about the field they have to operate in.
This will most likely come in the form of improved descriptive and robust feature representations for each plant, and focus on clustering methods capable of dealing with unbalanced data.

\section{Conclusion}\label{sec:conclusion}
This work presents a rapidly deployable weed classification system using visual data for enabling autonomous precision weeding without assuming prior weed species knowledge.
We introduce a three stage pipeline consisting of initial field surveillance, offline processing and data labelling, and precision weeding with which the system could be practically implemented.
We provide details for each stage and show the results of testing several offline processing and data labelling procedures.
Testing on field data and a larger leaf dataset, we perform a thorough analysis of these techniques for our classification system.
We demonstrate that the proposed system can label 12.3 and 23.3 times fewer samples than traditional full data labelling whilst achieving 78\% and 97\% classification accuracy for each respective dataset, without any prior knowledge of weed species before deployment.

\section*{Acknowledgements}
We would like to acknowledge the contributions of the Grains Research and Development Corporation (GRDC) for providing some of the funding for this work.

{\small
	\bibliographystyle{IEEEtran}
	\bibliography{refs}
}
\end{document}